# Computational Anatomy for Multi-Organ Analysis in Medical Imaging: A Review

*Juan J. Cerrolaza* [a*], *Mirella López Picazo* [b, c], *Ludovic Humbert* [c], *Yoshinobu Sato* [d], *Daniel Rueckert* [a], *Miguel Ángel González Ballester* [b,e], *Marius George Linguraru* [f, g]

[a] *Biomedical Image Analysis Group, Imperial College London, United Kingdom*
[b] *BCN Medtech, Universitat Pompeu Fabra, Barcelona, Spain*
[c] *Galgo Medical S.L., Spain*
[d] *Graduate School of Information Science, Nara Institute of Science and Technology (NAIST), Nara, Japan*
[e] *Institució Catalana de Recerca i Estudis Avancats (ICREA), Barcelona, Spain*
[f] *Sheickh Zayed Institute for Pediatric Surgical Innovation, Children's National Health System, Washington DC, USA*
[g] *School of Medicine and Health Sciences, George Washington University, Washington DC, USA*

## Abstract

The medical image analysis field has traditionally been focused on the development of organ-, and disease-specific methods. Recently, the interest in the development of more comprehensive computational anatomical models has grown, leading to the creation of multi-organ models. Multi-organ approaches, unlike traditional organ-specific strategies, incorporate inter-organ relations into the model, thus leading to a more accurate representation of the complex human anatomy. Inter-organ relations are not only spatial, but also functional and physiological. Over the years, the strategies proposed to efficiently model multi-organ structures have evolved from the simple global modeling, to more sophisticated approaches such as sequential, hierarchical, or machine learning-based models. In this paper, we present a review of the state of the art on multi-organ analysis and associated computation anatomy methodology. The manuscript follows a methodology-based classification of the different techniques available for the analysis of multi-organs and multi-anatomical structures, from techniques using point distribution models to the most recent deep learning-based approaches. With more than 300 papers included in this review, we reflect on the trends and challenges of the field of computational anatomy, the particularities of each anatomical region, and the potential of multi-organ analysis to increase the impact of medical imaging applications on the future of healthcare.

## 1. Introduction

Organs in the human body are organized in complex structures closely related to their function. However, conditioned by the technological limitations of the moment, such as computational capacity and image resolution, and the limited availability of data, single-organ-based models are frequently used as an oversimplification of the complex human anatomy. Human organs are not only spatially, but also functionally and physically interrelated. In fact, these inter-organ relations are frequently exploited by radiologists when navigating and interpreting medical images. For example, in the absence of visually recognizable features in the image (e.g., characteristic intensity levels or texture patterns), radiologists naturally incorporate topographical anatomical knowledge to distinguish some structures from their surrounding tissues and neighboring organs (Sheridan and Reingold, 2017; Swensson, 1980; Telford and Vattoth, 2014). This intuitive use of contextual information has inspired numerous algorithms, proving effective for the location and segmentation of some of the most challenging organs, such as the pancreas (Chu et al., 2013; Hammon et al., 2013; Shimizu et al., 2010). The automatic interpretation of medical images can significantly benefit from the comprehensive analysis of multiple organs through computational anatomy, which is key to a wide range of applications, including diagnosis and therapeutic assistance (Kobatake, 2007), radiotherapy planning (Fritscher et al., 2014; Hamacher and Küfer, 2002; Hensel et al., 2007; Kaus et al., 2007; Qatarneh et al., 2003), surgery simulation (Si and Heng, 2017), or injury severity prediction (Hayes et al., 2013).

Although the number of publications on multi-organ analysis has traditionally been lower than those focused on a single organ, the development of more holistic and anatomically accurate approaches has been a constant since the early years of medical image analysis. In recent years, there has been an exponential increase in the number of papers on this topic. Fig. 1 shows the histogram of the papers introducing novel multi-organ analysis techniques cited in this review. Not coincidentally, the growing interest in computational anatomy techniques and multi-organ analysis occurs in parallel to the systematic increase in the number of imaging examinations worldwide. As a reference, Fig. 1 also shows the evolution in the number of CT and MR scans in England (NHS, 2013), observing similar trends throughout the world. The availably of large imaging datasets, together with the continuous increase in computing power, have been instrumental to the progress of medical image analysis in general, and to the development of efficient and sophisticated approaches to model inter-organ interactions in particular.

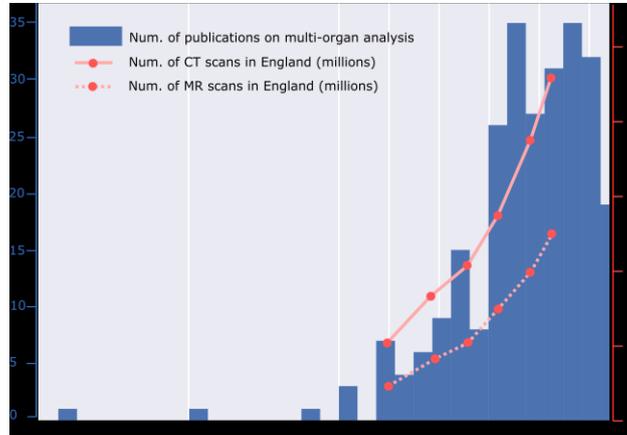

Fig. 1. Publications on multi-organ analysis (included in this review) per year (in blue) vs. the number of annual CT (pink solid line) and MR (pink dotted line) scans in England between 1995 and 2013 (NHS, 2013).

The modeling and analysis of multiple objects has always raised great interest in the computer vision field, with several reviews available on the subject (Leal-Taixé et al., 2017; Luo et al., 2014). However, despite the close relation between both fields, the analysis of multiple anatomical structures requires the development of new approaches specifically tailored to the context of medical imaging (e.g., the use of 3D images in different modalities, typically affected by noise, artifacts, and low contrast, or the need for accurately segmented and anatomically consistent results). From the simpler strategies based on global statistical models, to the more sophisticated sequential, or multi-level models, different strategies have been proposed to address the significant additional challenges involved in modeling multiple anatomical structures. These challenges include the simultaneous characterization of the inter-organ relations together with the particular locality of each organ, the use of complex anatomical and pose priors, or the need for geometrical constraints that prevent overlapping between organs, among others.

This paper presents, for the first time, a detailed review of the existing literature on computational anatomy techniques for multi-organ analysis. With more than 300 papers included in this survey, and with a special focus on the methodological aspects of those works, we propose a structured analysis of the different approaches commonly used when working with multi-organ complexes.[1] The paper is organized as follows. Section 2 presents an overview of the multi-organ parameterization models in the context of shape analysis, a prominent approach in computational anatomy. Section 3 discusses in detail the different methodologies commonly used in this field. To do so, we use the following categorization: global and individual models (Section 3.1), coupled deformable models (Section 3.2), multi-level models (Section 3.3), sequential models (Section 3.4), atlas-based models, (Section 3.5) machine learning-based models (Section 3.6), graph-based models (Section 3.7), and articulated models (Section 3.8). In Section 4, we discuss the trends and challenges that are specific to different anatomical region (e.g., abdomen, head, chest, etc). (Section 4.1), as well as the general limitations, challenges, and future trends in computational anatomy and multi-organ analysis (Section 4.2). Conclusions are presented in Section 5.

## 2. Parameterization of Multi-Organ Complexes

Organ parameterization is a fundamental problem in computational anatomy, and particularly relevant in the analysis of multi-organ structures. However, most of shape representations traditionally used, such as landmarks (Bookstein, 1991; Cootes et al., 1995; Dryden and Mardia, 1998), medial models (Blum, 1973; Pizer et al., 2000), moment invariants (Poupon et al., 1998), implicit representations (Leventon et al., 2000), and parametric representations (Brechbühler et al., 1995; Gerig et al., 2001; Staib and Duncan, 1992) were originally developed for single-organ applications, parameterizing the intrinsic shape of each structure separately, and thus neglecting the inter-structural shape correlation among organs. Moreover, the analysis of a multi-organ complex may be adversely affected by the independent parameterization of each organ, artificially increasing the entropy of the system, or even inducing anatomical inconsistencies such as collisions between objects, which should be later corrected by imposing restrictions to the model (Paragios and Deriche, 2000; Samson et al., 2000). In the following paragraphs, we present the types of parameterizing techniques that have been used in multi-organ analysis.

---

[1] In anatomy, the definition of organ can be rather vague or ambiguous (Wakuri, 1991). Here, we use the terms "organ" and "structure" interchangeably to refer to an anatomical region composed of one or several tissues that occupy a particular position and with a definite shape and function that are present independently in the body (e.g., liver, pancreas, heart, brain), or as part of a larger anatomical entity (e.g., subcortical structures of the brain).

Landmark-based representation is one of the most popular parameterization techniques, thanks to its simplicity and ease to deal with multiple objects. However, the quality of the landmarks directly affects the statistical efficiency of the resulting shape model, particularly critical in a complex multi-organ scenario. In (Duta and Sonka, 1998), the authors modeled neighboring structures in 2D images by defining common landmarks for adjacent regions. Despite its simplicity, this approach is also rather limited, only applicable to contiguous structures in direct contact with each other. Frangi et al. (Frangi et al., 2002) proposed a more general framework to define dense landmark correspondences in a multi-organ context, using non-rigid registration to propagate the landmarks from a multi-label atlas. Alternatively, Cates et al. (Cates et al., 2008) presented the landmarking of multiple objects as an optimization problem that minimizes a combined entropy-based cost function to define optimal surface point correspondences.

In a different type of approach, Pizer et al. (Pizer et al., 2000) used m-reps, an extension of the original medial axis descriptor (Blum, 1973) including anatomical and topological constraints. The extension to a more complex multi-organ scenario was later explored in several works (Fletcher et al., 2002; Pizer et al., 2003). However, the use of a single-organ-based parameterization remains the norm, using the inherent multi-resolution nature of m-reps to impose higher-level inter-organ constraints (see Section 3.3.3, and Section 3.4.2.3).

Level set functions (Samson et al., 2000) are another popular family of implicit representations of single-organ shapes, and one of the preferred methods to model multi-organ complexes via coupled deformable models (see Section 3.2). Most works use a simple, but computationally inefficient solution, in which each organ is modeled by a separate level set function. Vese and Chan (Vese and Chan, 2002) proposed a multi-phase parameterization able to represent $N$ objects with only $log_2(N)$ level sets using combination rules. Their approach prevented overlaps and gaps between the objects. In (Fan et al., 2008b), the authors presented an alternative parameterization using four level set functions to segment any number of objects, while the methods proposed in (Holtzman-Gazit et al., 2003) or Jeon et al. (Jeon et al., 2005) required only one function. All of these approaches are actually variations of the original multiphase framework presented by Vese and Chan (Vese and Chan, 2002), and their application is restricted to nested anatomical structures (e.g., white and gray matter in the brain).

The type of parameterization used and its potential for generalization to a multi-object context are closely linked to the multi-organ analysis framework adopted. For instance, landmark- or medial-based models are particularly suitable for hierarchical, sequential, or articulated approaches, while the use of implicit representation, or simple label-based binary masks are more adequate for coupled, or machine learning-based models, respectively. A detailed description of these multi-organ analysis strategies, variations, advantages and drawbacks are discussed in the next section.

## 3. Methodological Approaches for Multi-Organ Analysis

The aim of this section is to provide a structured reference guide for the different techniques of multi-organ analysis from a methodological point of view. While some of these methods were originally developed for a particular anatomical region, or clinical application, all these works share a common goal: to efficiently model the complex multi-organ human anatomy. Therefore, many of these techniques can be applicable to different anatomical contexts (see Section 4 for a more region-oriented discussion). Based on the papers surveyed in this review, we identified eight major categories and several other sub-categories of methods: global and individual models, coupled deformable models, multi-level models, sequential models, atlas-based models, machine learning-based models, graph-based models, and articulated models.

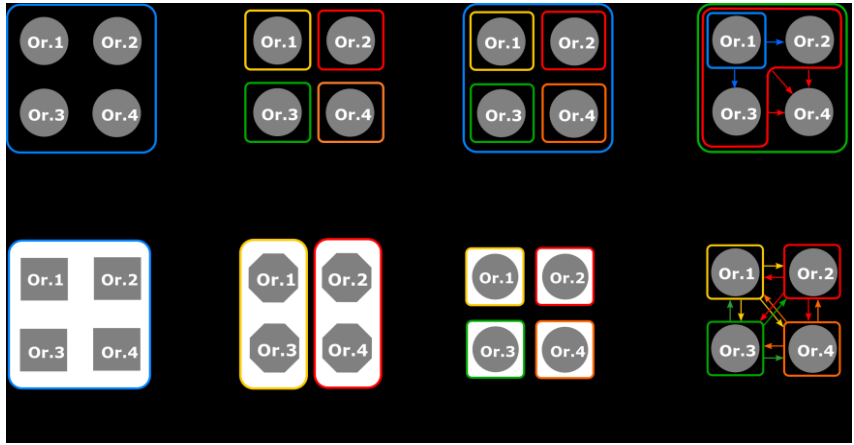

Fig. 2. Multi-organ modeling strategies. Squares, octagons, and circles indicate low, intermediate, and high resolution representation of the organs, respectively. Each color box indicates a separate statistical model. (a) Global model: all the organs (Or.) are modeled together as a single structure. (b) Individual models: each organ is modeled separately. (c) Nested models: inter- and intra-organ variability is modeled using a multi-level strategy, combining global and individual models. (d) Sequential models: organs are modeled sequentially according to a decreasing order of stability. Or.1 is used to estimate Or.2 and Or.3. Or.4 is finally modeled based on the anatomical constraints imposing by the former organs. The arrows indicate conditional relations between organs. (e) Multi-resolution models: global inter-organ constraints are imposed at coarser resolutions. Finer inter- and intra-organ details are modeled at higher resolutions. (f) Fully connected model: each organ-specific model includes the relations with all the surrounding organs.

## 3.1 Global and Individual Models

Some of the early works on multi-organ analysis conceived this new scenario as a particular case of traditional single-organ frameworks. The multi-organ complex was thus considered as a single object and tackled by using a global statistical model (Fig. 2(a)). In this context, the point distribution model (PDM) introduced by Cootes et al. (Cootes et al., 1995) has been one of the most extended and popular frameworks thanks to the versatility of its landmark-based parameterization. Landmarks of different organs were simply concatenated in the same vector, i.e. a single PDM was used to represent both the peculiarities of each subpart of the complex assembly as well as the spatial relations between them via principal component analysis (PCA). This approach has been extensively exploited by early (Duta and Sonka, 1998; Frangi et al., 2002; Fripp et al., 2007; Smyth et al., 1997; van Ginneken et al., 2006), and more recent works (Li et al., 2016; Mansoor et al., 2017; Picazo et al., 2018; Schwarz et al., 2010), and has been effective applied to different anatomical contexts and image modalities. However, despite the popularity of PDM to generate single- and multi-organ shape models, linear PCA is limited to the representation of data lying on a Euclidean vector space. Principal geodesic analysis (PGA) is a non-linear generalization of PCA introduced by Fletcher et al. (Fletcher et al., 2004) for modeling the variability of data lying on a manifold. This development is relevant to inter-organ relations on landmark-based parameterizations (e.g., relative pose or rotations) (Bossa et al., 2011; Bossa and Olmos, 2007, 2006), as well as alternative non-landmark-based parameterizations (e.g., m-reps) (Gorczowski et al., 2010; Styner et al., 2006, 2003), since they define a non-linear transformation space, and should be modeled as elements of a Riemannian manifold.

The advantages of integrating multiple organs into a global model are the computational simplicity of the approach, and the ability to overcome the weak image features that may be present in parts of multi-organ complex. The use of a global model can thus help to impose strong anatomical constraints that may facilitate the definition of those parts suffering from missing data, occlusion, or image artifacts (e.g., using part of the right ventricle and the left atrium of the heart to improve the segmentation of the left ventricle in echocardiogram images (Cootes et al., 1995, 1994)). On the other hand, global models are often inflexible and can be severely affected by the high-dimension-low-sample-size (HDLSS) problem (i.e. the dimensionality of the problem is significantly higher than the number of training images available), of particular relevance in a multi-organ context (Jung and Marron, 2009). Moreover, global models do not represent the scale of the organ, which limits their ability to accurately characterize the local geometry of organs.

As opposed to global models, some authors have adopted an equally simple approach, modeling each organ individually (Asl and Soltanian-Zadeh, 2008; Bagci et al., 2012) (Fig. 2(b)). Such models often require the pre-alignment of the training set, projecting the multi-organ complex to a common normalized shape space to guarantee the anatomical coherence, and to preserve the inter-organ spatial relations (Bagci et al., 2012; Yao and Summers, 2009). This single-organ-based strategy mitigates the HDLSS problem by simply reducing the dimensionality of each statistical model, while sacrificing their capacity to represent high-level anatomical patterns.

Global and individual organ models have been typically compared for their ability to quantify paired cardiac structures (Schwarz et al., 2010), or moderately variable brain regions (Akhoundi-Asl and Soltanian-Zadeh, 2007). All of these studies demonstrated that even the high level of global modeling of multi-organs is advantageous over individually modelling each organ separately. More recent studies have also shown that the co-modeling of organs like liver and spleen, known to be highly variable and not directly connected, can improve the segmentation of these organs as compared to individual models (Gollmer et al., 2012). However, these simple global models do not embed any inter-organ relations; more sophisticated models inspired from anatomy and physiology are presented in the following sections.

## 3.2 Coupled Deformable Models

The use of deformable models has been extensively studied in a variety of applications, including segmentation, tracking, and morphological analysis of organs (McInerney and Terzopoulos, 1996). Their ability to integrate (bottom-up) constraints derived from the image data with some (top-down) a priori knowledge about the location, size, and shape of the target structure is particularly interesting in the context of medical image analysis, and has led to some of the most promising single-organ segmentation methods (Ghose et al., 2012; Heimann et al., 2009). While deformable models are still in the core of some of the most sophisticated multi-organ analysis frameworks (discussed in subsequent sections), early multi-organ frameworks used traditional deformable models, sometimes as mere extension of their single-organ methods (Brox and Weickert, 2004), or imposing additional forces that prevent overlapping, gaps, or topology changes between organs (Namías et al., 2016; Zimmer and Olivo-Marin, 2005).

The level set framework (Samson et al., 2000) has been the preferred method to approach the multi-organ extension of deformable models (Brox and Weickert, 2004; Fan et al., 2008b; Kohlberger et al., 2011, 2007; Paragios and Deriche, 2000; Pohl et al., 2007; Samson et al., 2000; Tsai et al., 2001; Uzunbas et al., 2013; Vese and Chan, 2002; Zimmer and Olivo-Marin, 2005), with only a few works exploring the use of explicit models (or "snakes" (Kass et al., 1988)) (Costa et al., 2007; Fang et al., 2011; MacDonald et al., 1994; Srinark and Kambhamettu, 2006; Zhu et al., 1995). In its most basic and popular formulation, each organ is associated with one level set function, an idea initially suggested by Zhao et al. (Zhao et al., 1996). This simple use of a separate contour for each organ is prone to produce intersections between contours, gaps, or topological changes, with little or no anatomical sense. Paragios and Deriche (Paragios and Deriche, 2000) and Samson et al. (Samson et al., 2000) were some of the first to use non-overlapping constraints to generate mutually exclusive propagating curves. Inspired by these works, Rousson et al. (Rousson et al., 2005) presented a Bayesian formulation for the segmentation of the prostate and bladder via coupled surfaces evolution, incorporating constraining terms to penalize voxels with multiple labels. Alternatively, Merino-Caviedes et al. (Merino-Caviedes et al., 2010) and Ma et al. (Ma et al., 2013) used the region competition speed term proposed by Brox and Weickert (Brox and Weickert, 2004), to segment the brain and the female pelvic region, respectively. The use of competing forces was also used by Yan et al. (Yan et al., 2009), as the strategy to deal with the gaps between subcortical structures. Also on the segmentation of subcortical structures of the brain, Kim et al. (Kim et al., 2014) proposed a semi-automatic framework in which the adjacent structures are iteratively segmented and corrected using repulsive forces derived from the previous iterations of the algorithm. Gao et al. (Gao et al., 2017, 2011) presented an interesting multi-organ segmentation framework based on Newtonian mechanics theory in which the interactions between contours are governed by the action/reaction principle, and thus preventing overlap. Finally, Ho and Shi (Ho and Shi, 2004) prevented overlap by incorporating prior knowledge of the rough spatial location of the structures, and thus confining the evolution of each level set function within a designated region of interest.

In some anatomical contexts, there may be prior knowledge about the relative distance between contours (e.g., epicardium and endocardium of the left hearth ventricle (Schwarz et al., 2010)). Thus the previous non-overlapping constraints can be replaced with distance-based coupled forces specifically tailored to the anatomical regions under study. Early works on distance-based coupled models focused on the segmentation of nested structures with a nearly constant thickness, such as the cerebral cortex (L. Wang et al., 2013; Zeng et al., 1999), or the myocardium (Kohlberger et al.,

2007). However, its extension to non-nested structures with more complex inter-organ distances is not clear. Li et al. (Li et al., 2005) presented a more general method to model the inter-organ space using graph theory. In the proposed framework, the authors combined level set functions with an explicit representation via triangulated meshes, modeling the inter-organ space using geometric graphs. Distance and geometrical constraints were encoded in the graph through the cost terms associated to the nodes and edges, using graph-cut algorithms to solve it. Thanks to the versatility of the model, the original framework was adapted by several authors to deal with different anatomical structures, such as the ankle bones (Li et al., 2005), the femur and tibia, or the bladder and the prostate (Song et al., 2013), and inspired others to create similar graph-based coupled models to segment the hip (Kainmüeller et al., 2009b) or the head bones (Kainmueller et al., 2009).

Alternatively to simple distance-based boundary condition, some authors (Brock et al., 2005; Hensel et al., 2007) incorporated organ-specific biomechanical properties and surface interfaces into a multi-organ finite element model. Moreover, as shown in (Brock et al., 2005) the use of tissue properties for neighboring organs (e.g., bladder and rectum) can also help to impose anatomically consistent constraint to the deformation of more challenging anatomical regions (e.g., the prostate).

The co-dependencies frequently observed between adjacent anatomical organs was also exploited by many authors to incorporate additional coupled forces into deformable-based segmentation models. In this line, Yang et al. (Yang et al., 2004) and Uzumbas et al. (Uzunbas et al., 2010) proposed similar approaches using the level set framework to formulate a maximum a posteriori segmentation model combining low-level intensity information with high-level shape priors from local neighboring organs via joint probability density functions. The incorporation of shape priors was also explored by Litvin and Karl (Litvin and Karl, 2005) who encoded high-level constraints in the form of cumulative distribution functions of single- (curvature and inter-node distance) and multi-organ (inter-organ distance) shape features. Alternatively to these hand-crafted shape features, Pang et al. (Pang et al., 2015) proposed a new dissimilarity metric to directly encode the difference between each organ and the corresponding training samples. By modeling the joint probability density function of adjacent structures, the proposed coupled model integrates organ-specific constraints as well as their relative position.

The ability to automatically handle topological changes is another property of many level set-based deformable models, particularly appreciated in the computer vision field. However, such flexibility can be disadvantageous in contexts in which the objects have a well-defined topology that must be preserved, as is often the case with anatomical structures. Mangin et al. (Mangin et al., 1995) presented a pioneer multi-organ, topology preserving, evolution framework that combines a sequence of homotopic morphological filters with a constrained model of the target anatomy. The method was later expanded by Bazin and Pham (Bazin and Pham, 2006) who generalized the model to non-concentric spherical structures. In (Fan et al., 2008a), Fan et al. integrated digital homeomorphism constraints into the level set formulation to preserve organ-topology and inter-organ relations. Han et al. (Han et al., 2003, 2002) presented a simple multi-organ deformable model robust to topological changes, introducing a digital point criterion to guarantee topological preservation during the evolution of the contours. However, the proposed method only incorporates mechanisms to avoid the merging of disjoint sets, applied in contexts with limited number of organs (e.g., segmentation of left and right ventricle in cardiac MRI (Arrieta et al., 2017)). Typically, the analysis of anatomical groups with higher number of organs requires the use of more comprehensive anatomical models able to represent the complex inter-organ relations and inter-organ variations. These approaches includes sequential, articulated, or multi-level models, discussed below.

## 3.3 Multi-Level Models

Inspired by the hierarchical nature of the human visual system, (Hubel, 1988; Marr, 1982), multi-level models decompose the data into different levels of detail according to a coarse-to-fine analysis rule. In a multi-organ context, the particular locality and details of each organ is modeled at finer resolutions (i.e., local scale), while broader inter-organ relations are considered at coarser levels (i.e., global scale). This representation of the information has offered an interesting new approach to multi-organ analysis, combining the robustness and specificity of global models with the flexibility and generality of single-organ-based strategies. There are several families of multi-level or hierarchical model used in computational anatomy, as detailed in the following sub-sections: nested statistical models, multi-resolution shape models, and medial models.

### 3.3.1 Nested Statistical Models

At the intersection of the global and individual models discussed in Section 3.1, one of the simplest multi-level approaches is the nested combination of multiple statistical models (Fig. 2(c)). At a lower level, each structural element (i.e., the smaller unit of information in which the anatomical complex is divided) is modeled independently by separate statistical models, for instance using PCA. Topological constraints are imposed by a higher level model, which typically operates over the parameterization obtained by the separate models to improve efficiency and alleviate the HDLSS problem. The use of nested statistical models was initially proposed by Bernard et al. (Bernard et al., 2001), using statistical parametrical models via PCA to segment and analyze multiple vertebrae in cervical spine X-ray images. Similarly, Zhang et al. (Zhang et al., 2011) proposed a two-levels PCA-based hierarchical model of multiple brain structures. However, only the center of mass was used in the global model, thus limiting its capability to characterize complex inter-organ relations. Bukovec et al. (Bukovec et al., 2011) presented a more general and flexible approach, creating a three-levels framework that combines PCA-based shape and appearance models. In this latter approach, higher-level topological models operated on the parameterization of previous models.

Yokota et al. (Yokota et al., 2009) explored the potential of multiple statistical shape models to represent adjacent articulated structures, such as the hip joint of the femur and the pelvis, combining global, inter-organ and organ-specific shape models. Unlike the previous nested approaches, all the models operate on the same landmark-based representation, using additional constraint terms to guarantee the correspondence between the nodes of the shapes generated by each model. Similarly, Okada et al.

(Okada et al., 2008b) presented a patch-based decomposition of the multi-organ complex, using an adhesiveness constraint to combine inter- and intra-organ patch shape models.

Aiming to overcome the inherent limitations of global models, Lecron et al. (Lecron et al., 2012a) proposed the use of multi-level component analysis (MLCA), a generalization of the popular PCA for analyzing multi-group or multi-set data (Timmerman, 2006) (i.e., data that can be divided into conceptually meaningful blocks). Unlike classic PCA, MLCA creates different sub-models for different blocks of information, allowing to analyze the within-block (i.e., localities) and between-block (i.e., global changes) variation separately. The flexibility of this multi-level-based model has been explored by several authors to analyze the vertebral body (Lecron et al., 2012a, 2012b; Neubert et al., 2014), where each vertebra represents a block of information. The potential of MLCA to model multi-organ structures was also used by Lee et al. (Lee et al., 2016) to create a hybrid multi-object model-based multi-atlas segmentation method for rodent brains.

Although nested models propagate the statistical information about organs through multiple levels (from organ to organ complex), they simply link the individual organ models to the global complex model through a function. In the next sub-section, we describe an essential advancement in the field of computational anatomy, the development of multi-resolution multi-organ shape models that provide continuity and coherence between the levels of representation.

### 3.3.2 Multi-Resolution Shape models

Davatzikos et al. (Davatzikos et al., 2003) proposed one of the first multi-resolution approaches for shape analysis combining the classic PDM framework with the wavelet transform. Initially conceived for 2D single-organ structures, the method was later extended to the multi-organ context by Cerrolaza et al. (Cerrolaza et al., 2011). The authors used the wavelet transform to decompose the objects into smaller blocks of information, each one being independently modeled via PCA. The lower-frequency blocks contain global information that guarantee the coherence of the general structure

(e.g., the general layout of the organs), while the higher-frequency components model the finer details. However, while this representation reduces the dimensionality of the model, thus alleviating the HDLSS problem, it is difficult to establish a direct relation between these frequency-based blocks and the graphical representation of the anatomical structures, which significantly reduces the robustness to image artifacts and noise. The potential of multi-resolution partitioned models in single-organ shape analysis was also exploited in previous works (Hontani et al., 2013; Okada et al., 2008a; Pereañez et al., 2015; Reyes et al., 2010, 2009; Yang et al., 2014; Zhao, 2006).

Based on the multi-resolution analysis framework originally proposed by by Lounsbery et al. (Lounsbery et al., 1997), Cerrolaza et al. (Cerrolaza et al., 2012) presented an intuitive cluster-based multi-resolution shape model where different groups of organs were modeled together at different scales (Fig. 2(e)). Unlike the original model presented by Davatzikos et al. (Davatzikos et al., 2003) based on the independent modeling of disjointed blocks of information, this new framework creates specific statistical shape models to explicitly characterize different inter-object associations at each scale (Cerrolaza et al., 2013, 2012). However, the hierarchical configuration was manually defined by the user, thus hindering its practical application when working with complex data and large numbers of organs. The framework was later extended to include a new agglomerative landmark clustering method that automates the hierarchical configuration of the algorithm (Cerrolaza et al., 2015, 2014). Additionally, the original method (Cerrolaza et al., 2012) was also extended to include the subdivision of organs into smaller anatomically significant sub-parts. This hierarchical model was also exploited by interactive annotation tools (Valenzuela et al., 2015) yielding a substantial speed-up on segmentations process.

Shen et al. (Shen et al., 2001) proposed an interesting alternative multi-resolution shape descriptor for triangle surfaces. Using the normalized volume of the tetrahedron formed by neighboring vertices at different distances, they defined an attribute vector able to describe the morphology of an object at different scales. However, the model lacks a global model to prevent the overlap between organs and to guarantee the anatomical coherence of the complex.

*3.3.3 Medial Models*

As mentioned in Section 2, m-reps-based have been successfully used to represent single-organ anatomical models (e.g. the kidney (Joshi et al., 2001)), using PGA (Csernansky et al., 1998; Fletcher et al., 2004; Styner et al., 2006). However, a hierarchical multi-figure approach (the term "figure" represents each of the medial sheets in which an object is decomposed) is needed to model in more detail structures such as the liver, the renal pelvis (Han et al., 2005), or the heart (Hui Sun et al., 2008). These multi-figure approaches (Han et al., 2005; Hui Sun et al., 2008) describe an object at successively smaller scales, following a coarse-to-fine hierarchy, thus enabling not only the analysis of the organ as a whole, but also of each individual figure within the organ and the relations among them.

The modeling of multi-organ structures has also been integrated into the m-reps framework by adding an upper scale level for the multi-object complex, which can be defined as the assembly of individual objects. For example, Pizer et al. (Pizer et al., 2003) presented one of the first multi-organ extensions of m-reps by simply computing global statistics on all the organs taken together; on the other hand, Vera et al. (Vera et al., 2012) combined individual medial models for the analysis of multiple abdominal models (similarly to the global and individual models discussed in Section 3.1). Later, Jeong et al. (Jeong et al., 2006) incorporated inter-organ neighboring effects to the model. In this framework, the variation of an organ was decomposed into two parts: self- and neighbor-effects. The former describes the variation of the organ itself; while the latter models the inter-organ interactions as a function of the neighbor's geometric descriptors using native m-reps operations (e.g., addition and subtraction of medial atoms), whose probability densities are estimated via PGA. In the work presented by Lu et al. (Lu et al., 2007), the authors exploited the multi-resolution properties of m-reps-based models to create a complete multi-scale multi-object framework. In the proposed framework, the deformations were divided into different levels of detail, describing smaller scale features (e.g., local details) as residues of larger scale features (e.g., object-based, or global multi-object deformations). The method is closely related to the sequential models discussed in Section 3.4, using Markov random field (MRF) models to characterize sequentially the statistical inter-object relations.

Medial models have been used successfully to model different multi-organ anatomical structures, such as the pelvis (Chaney et al., 2013; Jeong et al., 2006; Lu et al., 2007), the bladder-rectum-prostate complex (Lu et al., 2007; Merck et al., 2008), or the hip musculoskeletal system (Gilles and Magnenat-Thalmann, 2010). However, despite their intrinsic multi-resolution nature, they may be less intuitive than alternative landmark-based representations, which are arguably the simplest and most popular methods used to represent shapes.

### 3.4 Sequential Models

In most of the works mentioned in the previous sections, inter-organ relations were implicitly characterized by combining multiple organs (or adjacent regions) into a common statistical model. Sequential models represent an alternative strategy where organs are analyzed consecutively following a predefined order of increasing complexity (Fig. 2(d)). The underlying hypothesis is that the analysis of the more challenging organs (e.g. of more complex and variable shape) can benefit from the analysis of more stable and related organs in the previous step. Some authors (Kéchichian et al., 2014) argue that simultaneous multi-organ analysis/segmentation approaches are better performers than sequential approaches because the latter require a segmentation sequence to follow, which also raise questions about how to avoid the propagation of errors through the sequence. Nevertheless, sequential methods have proven to be very effective for the segmentation of some of the most challenging organs, such as the pancreas (Erdt et al., 2011; Hammon et al., 2013; Shimizu et al., 2010) or the gallbladder (Huang et al., 2014), using surrounding organs (e.g., liver and spleen) as support structures. Sequential approaches have also performed well for the identification of pathologies that may manifest as abnormalities in the anatomical arrangement of certain structures (e.g., brain tumor identification and segmentation (Batrancourt et al., 2006; Khotanlou et al., 2009; Puentes et al., 2008)).

Despite the hierarchical organization of the organs being intuitively linked to our perception of the human anatomy, and thus offering a unique advantage to the definition of sequential models, these approaches raise two important questions: 1) how to define the order in the analysis of the organs, and 2) how to parameterize the inter-organ dependencies or relations. The following sub-sections survey alternative approaches to answer the above questions.

### 3.4.1 Sequential Organization of Multiple Organs

The performance of sequential analysis methods for multi-organs is highly conditioned by the order in which the organs are analyzed. Based on the idea that stability provides efficiency, many works assume a predefined order of the organ analysis, usually defined heuristically, and driven by some intuitive notion of stability and/or inter-organ relations (Bloch et al., 2005, 2003; Camara et al., 2004; Colliot et al., 2006; Hudelot et al., 2008; Jeong et al., 2008). Some authors solved the problem from a more quantitative approach, using image-based stability criteria, such as contrast or intensity variability, to define the order in which the organs were processed (Fletcher et al., 2002; Lu et al., 2007; Pizer et al., 2005). The anatomy of the human body has also been frequently exploited (Bloch et al., 2005; Fasquel et al., 2006; He et al., 2015; Sun et al., 2016; Udupa et al., 2013, 2011; Udupa and Saha, 2003; Wang and Smedby, 2014a, 2014b, 2015), using inter-organ relations, such as proximity (e.g., liver and the pancreas (Erdt et al., 2011; Hammon et al., 2013; Shimizu et al., 2010)), symmetry (e.g., left and right kidney (Camara et al., 2004)), inclusion (e.g., thoracic cavity and lungs (Camara et al., 2004; Udupa et al., 2011; Wang and Smedby, 2014b)), or intersection (e.g., hepatic vessels intersecting the liver (Fasquel et al., 2006)) to improve the computational models. However, the definition of the optimal sequential order depends on the anatomical organs involved in the analysis, the imaging modality, and the mathematical analysis framework, and remains an active area of research.

The goal of optimal organ hierarchy that defines the sequence of the analysis is to determine, among all possible combinations, what sequence provide the best multi-organ segmentation (i.e., to maximize the overall segmentation accuracy), or organ prediction (i.e., to find the sequence of optimal predictors for a given missing organ). While a heuristic approach could be an option when the number of organs is small (Rousson and Xu, 2006), the number of all possible combinations becomes intractable with the increasing complexity of the analysis task. Rao et al. (Rao et al., 2008, 2006) proposed the use of multivariate statistical analysis tools, such as canonical correlation analysis (CCA), to establish optimal prediction paths for any pair of sub-cortical structures. They assumed an inverse relation between the inter-organ correlation strength and the prediction error of a target organ. Similarly, Okada et al. (Okada et al., 2015) combined CCA and partial least squares regression (PLSR) to create a directed inter-relation graph of abdominal organs. Interestingly, the proposed CCA-based correlation map confirmed some of the intuitive ideas about inter-organ relations initially exploited by early works on multi-organ analysis (e.g. (Camara et al., 2004; Udupa and Saha, 2003)), showing strong correlation patterns between structures with a proximity (e.g., left lateral ventricle and the left thalamus), or symmetry (e.g., left and right lateral ventricles) relation. These patterns of anatomical organization can be robustly represented using the mathematical structure of graphs (Atif et al., 2007).

Graphs provide a simple, yet compact representation of organs (nodes) and their relations (edges) from simple spatial relations (e.g., distance, orientation) to texture similarities (e.g., intensity profile differences (Matsumoto and Udupa, 2013)) [2]. In (Fouquier et al., 2007), Fouquier et al. used classic graph-based optimization algorithms to automatically establish the sequential segmentation path for graphs of brain structures. In particular, two types of algorithms were studied: edge- (e.g., shortest path or maximal flow) and path-based (minimal global entropy) optimization methods. However, only small graphs with four cerebral structures were considered in the study. The shortest-path algorithm was also used by Matsumoto and Udupa (Matsumoto and Udupa, 2013) to guide the computerized automatic anatomy recognition in the thorax, while Zhan et al. (Zhan et al., 2008) proposed to study the sequential detection and segmentation of the whole-body using information theory. The authors used conditional probability to model the dependency between tasks (i.e., detection, or segmentation of a particular organ), using information gain-based criteria to optimize the process. Alternatively, Fouquier et al. (Fouquier et al., 2012, 2008) explored a visual attention-based criteria to create an optimal segmentation sequence of brain structures by combining spatial information with saliency maps.

---

[2] *Here, graphs are considered as an abstraction of the multi-organ information, not necessarily linked to the corresponding image domain in which the organs are embedded. Therefore, the final segmentation relies on other methods, not necessarily on graphical segmentation methods. See Section 3.7 for graph-based models.*

Besides the definition of the sequence of actions in the modeling and analysis of multi-organs, the other fundamental element in the design of sequential architectures is the mathematical parameterization of the inter-organ relations, which is reviewed next.

### 3.4.2 Inter-organ Relations

In this section, we discuss the most relevant mathematical frameworks proposed in the literature to encode a wide range of inter-organ relations, from well-defined distance measurements, like the Euclidean distance between two organs, to fuzzier or intuitive concepts such as "close to", or "in the proximity of" (Hudelot et al., 2008). We grouped the inter-organs relations into four categories, based on the methodological approach: regions of interest, fuzzy connectedness, medial models, and multivariate linear statistical methods.

#### 3.4.2.1 Regions of Interest

Conditional regions of interest (ROIs) were some of the first representations used to integrate simple inter-organ spatial relations, such as topology, distance, and orientation (Gapp, 1994; Kuipers and Levitt, 1988) into a multi-organ analysis framework. In (Camara et al., 2004), Camara et al. presented a sequential strategy for the segmentation of thoracic and abdominal structures, using ROIs to express spatial constraints inferred from previously segmented structures. Using a predefined sequence of organs, the segmentation of each organ was confined to a ROI defined with respect to the previous organ in the sequence based on simple inclusion (the lungs are inside the skeleton, and the skeleton inside the skin), or proximity (e.g., the lungs and the kidneys) relations. Similarly, ROIs were also used by Ma et al. (Ma et al., 2010) to impose spatial and topological restrictions in the shape of adjacent structures (e.g., muscles of the pelvic floor), and by Fasquel et al. (Fasquel et al., 2006) to improve robustness and processing efficiency in an general-purpose interactive sequential segmentation tool.

*3.4.2.2 Fuzzy Connectedness*

The fuzzy treatment of the geometric and topological relations among organs can be considered as a generalization of the more traditional rule-based definition of the ROI, often relying on rigid and predefined propositional logics. However, despite the intuitive notion of spatial connection between organs, these relations can be vague and diffuse in practice. A limiting factor is the heterogeneity of the organs' intensity values as a consequence of the spatial and temporal resolution of imaging devices. The use of fuzzy sets as representation framework of spatial relations was early suggested by Udupa and Saha (Udupa and Saha, 2003), and later exploited by Geraud et al. (Géraud et al., 1999), Bloch et al. (Bloch et al., 2005, 2003) or Colliot et al. (Colliot et al., 2006) in the sequential segmentation of different components of the brain. The authors used spatial fuzzy sets to combine organ-level information (e.g., shape- or intensity-based features) with inter-organ constraints (e.g., "inclusion in" or "exclusion from" organs previously segmented, relative distance, or directional position). Fuzzy sets also provide a natural framework for the mathematical representation of more uncertain and intuitive spatial concepts (e.g., "near to", "on", or "between"), establishing a direct between these abstract concepts of natural language and the quantifiable information extracted from the image. In this context, Hudelot et al. (Hudelot et al., 2008) used fuzzy sets to create an ontology of spatial inter-organ relations, exploiting user-supervised prior knowledge of the spatial organization of the structures (often expressed in linguistic terms) to guide the interpretation, recognition and analysis of radiological images.

Besides spatial relations, several authors (Matsumoto and Udupa, 2013; Sun et al., 2016; Udupa et al., 2013, 2011) also explored the used of fuzzy models to encode shape uncertainties in the image space. Similarly to a probabilistic atlas, the proposed fuzzy organ model (Udupa et al., 2013, 2011) was defined as a membership function that represents the degree of association of each voxel to the specific organ, computed as the normalized average of a set of pre-aligned distance maps. These fuzzy organ models were later combined with spatial parent-to-child relations between organs derived from a pre-defined hierarchy of thoracic (Matsumoto and Udupa, 2013; Sun et al., 2016; Udupa et al., 2013, 2011) and abdominal organs (Udupa et al., 2013).

### *3.4.2.3  Medial Models*

In addition to its multi-resolution nature (see Section 3.3.3), the m-reps representation can also establish sequential inter-organ relations by defining neighboring atoms. Neighboring atoms define the location of one figure in the object-intrinsic coordinate system of a neighboring organ. This property was used by Fletcher et al. (Fletcher et al., 2002) to create a hybrid multi-resolution sequential model of the pelvic region. In the proposed framework, global multi-organ deformations were controlled at a higher resolution level, while each organ was later refined at a lower resolution level using predictions from previously segmented organs. In the work presented by Pizer et al. (Pizer et al., 2005), these neighboring atoms were defined based on simple proximity relations between the target organ and the adjacent structures, and used to propagate shape deformations between them according to a predefined order of decreasing stability. In the proposed framework, the deformation of an organ was defined as the combination of sympathetic (i.e., deformations infered from previous organs in the analysis sequence) and residual changes (based on the residue idea presented by Lu et al. (Lu et al., 2007)), using PGA to estimate the corresponding probabilities. The framework was further developed by Jeong et al. (Jeong et al., 2008) using principal component regression to estimate the conditional mean and the shape distribution of the prostate from the m-reps representation of the bladder and rectum.

### *3.4.2.4  Multivariate Linear Statistical Methods*

Two multivariate linear statistical methods have been particularly used in the quantification and parameterization of inter-organ relations: CCA and PLSR.

Typically, CCA has been used to quantify the degree of correlation between pairs of organs (or even between specific regions of organs (Yokota et al., 2013)), and is at the core of many technique for automatic (or semi-automatic) optimization of sequential methods (see Section 3.4.1). The use of CCA as shape predictive tool was first proposed by Liu et al. (Liu et al., 2004) to estimate occluded brain regions (e.g., obscured by tumors) from visible adjacent structures. Also in brain imaging, Rao et al. (Rao et al., 2008, 2006) used CCA and PLSR to create a one-to-one correlation map between 18 subcortical structures, which they used to establish robust prediction sequences to estimate missing organs, and improve the segmentation of the more challenging organs. A similar approach was used by Okada et al. (Okada et al., 2012) in one of the first automatic hierarchical frameworks for the segmentation of multiple abdominal organs. In the proposed framework, the set of organs was initially divided in two categories ("stable" and "variable") according to the accuracy obtained by a preliminary segmentation. This categorization of organs was later combined with CCA to create an organ correlation graph, which established the sequence in which the organs were finally segmented. Shape constraints were also imposed combining single- and multi-organ statistical shape models (SSMs) (Okada et al., 2008a). This binary division of organs was later expanded by including a third "intermediate" category of organs (Okada et al., 2015, 2013), still dependent on the "stable" organs (e.g., the liver), but able to improve the analysis of the most challenging structures (e.g., gallbladder).

Weighted PCA was also used to encode a hierarchical organization in landmark-based multi-organ SSMs. Chandra et al. (Chandra et al., 2016) proposed the use of multiple weighted shape models to activate and deactivate organs according to an order of decreasing stability so that the most challenging organs are initialized by the most robust ones. Inspired by the early work of de Bruijne et al. on neighbor-conditinal shape models (de Bruijne et al., 2007), Cerrolaza et al. (Cerrolaza et al., 2016) presented a generalization of the organ correlation graph proposed by Okada et al. (Okada et al., 2012) using examples in brain and abdominal modeling. Unlike the classic organization of the information in the form of rigid predictor-target relations, the authors proposed to represent the inter-organ relations as a general, fully-connected network, in which all the organs are interrelated (Fig. 2(f)). This framework used a generalized PCA to generate a separate organ-specific SSM for each object that includes all its surrounding structures (see Fig. 2 (f)). The result was a set of weighted SSMs, where the organ-based weights encoded the degree of the relation of each organ with its neighbors. Using CCA to define the weights automatically, the authors also demonstrated how the proposed framework is a generalization of the classic landmark-based global (Cootes et al., 1995) and sequential approaches (Okada et al., 2012).

## 3.5 Atlas-based Models

Atlases are reference anatomical templates that provide prior anatomical knowledge, typically from manually annotated data, for medical image analysis (Iglesias and Sabuncu, 2015; Sanroma et al., 2016). Using multiple organ labels, atlas-based models encode shape, spatial locations, as well as spatial inter-organ relations. Brain imaging has been the prevalent field of application (Aljabar et al., 2009; Bazin and Pham, 2008; Christensen et al., 1997; Cocosco et al., 2003; Collins et al., 1995; Dawant et al., 1999; Fischl et al., 2002; Han and Fischl, 2007; Heckemann et al., 2006; Lancaster et al., 1997; Mazziotta et al., 2001, 1995; Rohlfing et al., 2004), thanks in part to the success of image registration techniques in this field, but atlases have also been extensively used for other anatomical regions, including the abdomen (Linguraru et al., 2009; Oda et al., 2012; Park et al., 2003; Schreibmann et al., 2014; Shimizu et al., 2007; Wolz et al., 2013), the head and neck (Han et al., 2008), pelvic structures (Akhondi-Asl et al., 2014; Isgum et al., 2009; Weisenfeld and Warfield, 2011), and skeletal muscle (Karlsson et al., 2015).

In the early days of atlas-guided medical analysis, atlases were formed by simple topological maps derived from one case (Talairach and Tournoux, 1988). In this simplest form, the multi-organ segmentation was treated as an image registration problem, and the inter-organ relations were deterministically defined by the reference map (Dawant et al., 1999; Evans et al., 1991; Sandor and Leahy, 1997). Typically, these single-atlas-based approaches relied on non-rigid deformation to account for the anatomical differences between subjects (Christensen et al., 1997; Collins et al., 1995; Dawant et al., 1999; Lancaster et al., 1997; Sandor and Leahy, 1997; Thompson et al., 2000). However, this approach is usually insufficient to model complex inter-organ relations and their variation among the population.

In a more sophisticated form, probabilistic atlases (Mazziotta et al., 1995) combine information from multiple observations to quantify human anatomical variability in the form of a probability map. Their potential to provide a statistical description of complex structures was exploited in the initial attempts at developing reference anatomical maps from a population (Chiavaras et al., 2001; Mazziotta et al., 1995; van Buren and Maccubbin, 1962). Typically, probabilistic atlases were built by registering all the examples to a common reference system. However, the registration can be problematic when including multiple organs with very different sizes, when the bigger organs (e.g., the liver in the case of the abdomen) drive the registration, which results in misaligning the smaller organs (e.g., the gallbladder). Other models adopted a most robust alternative approach using multi-resolution non-rigid registration (Chen et al., 2017a), registering each organ individually (Park et al., 2003), or performing efficient feature-based registration (Fejne et al., 2017). Despite this improvement, the direct integration of probabilistic atlases into a multi-organ segmentation framework (e.g., in combination with a maximum a posteriori approach (Liu et al., 2010; Okada et al., 2013; Park et al., 2003; Shimizu et al., 2007)) is still challenging due to the large variation in shape and inter-organ relations. In one approach, a spatially-divided atlas was proposed to deal with the large variability in local areas (Chu et al., 2013). In practice, probabilistic atlases are typically combined with additional contextual information, such as topological and shape priors (Bazin and Pham, 2008; Cocosco et al., 2003; Linguraru et al., 2010; Okada et al., 2012, 2008b; Pohl et al., 2006; Wang et al., 2012; Zhou and Bai, 2007), or combined with Bayesian Models (Gubern-Mérida et al., 2011; Park et al., 2003), or MRF (Fischl et al., 2002; Han and Fischl, 2007; Park et al., 2010; Van Leemput et al., 1999) in order to refine the boundaries between organs.

Multi-atlas based analysis (Rohlfing et al., 2003a, 2003b) is a special category of probabilistic atlases where each atlas in the training set is available and potentially used for analyzing the new image. In general, multi-atlases performed better than single-atlases for multi-organ segmentation (Fejne et al., 2017; Iglesias and Sabuncu, 2015). The possibility to optimize the selection of exemplars for each target image (Aljabar et al., 2009; Wolz et al., 2013; Xu et al., 2015), and the use of more sophisticated label fusion techniques, (including majority (Heckemann et al., 2006; Klein et al., 2005), weighted voting (Isgum et al., 2009; H. Wang et al., 2013), hierarchical modeling (Wolz et al., 2013), or probabilistic reasoning (Warfield et al., 2004)), allowed to create accurate patient-specific models that reduced the uncertainty in the inter-organ regions. The reader is referred to (Iglesias and Sabuncu, 2015; Sanroma et al., 2016) for a comprehensive survey of multi-atlas segmentation strategies and their applications.

*3.6 Machine Learning-Based Models*

Machine learning techniques have always played an important role in the medical imaging field (Wang and Summers, 2012; Wernick et al., 2010). However, it is recently that these techniques have raised an even higher interest in the research community, benefiting from the increase in computing processing power, and from the availability of large databases. For multi-organ analysis, machines can directly learn the global image context and inter-organ relations from examples, instead of explicitly modeling the dependency between structures based on algorithms.

Early machine learning works used a series of multiple organ-specific classifiers (Sofka et al., 2010; Zhan et al., 2008). Typically, the performance of these simple approaches was directly related to the number of organs in the study, and it was conditioned by the predefined sequence or hierarchical inter-organ relation, as discussed in Section 3.4.1. The succession of different classifiers at multiple locations and scales is slow, and does not scale up well as the number of organs increases. Alternatively, common classifiers that encourage feature sharing across classes (i.e., organs) provide a more efficient approach. Moreover, as the shapes of adjacent organs are often correlated, the appearance of neighboring image regions provides a valuable source of information in a multi-organ analysis scenario.

We organized the section into three main components of machine learning approaches: single-classifier, multi-classifier, and deep learning methods.

*3.6.1 Single-Classifier Models*

A single-classifier strategy provides good generalization, is computationally fast, and also learns relevant inter-class relations (Torralba et al., 2007). In this line of work, Criminisi et al. (Criminisi et al., 2009) presented a single random forest (RF)-based regressor for the simultaneous detection of multiple anatomical structures in the abdomen and thorax. Although only the organ centers were computed, the model was later refined in (Criminisi et al., 2013) to estimate the bounding box containing each organ. In these works, the authors introduced the use of context-rich visual features as an efficient strategy to capture inter and intra-organ spatial information. These long-range spatial features (i.e., pairwise comparison of mean intensities over displaced, asymmetric cuboidal regions of the volume) quickly became a popular set of hand-crafted features exploited by many machine learning-based multi-organ analysis methods (Fischer et al., 2014; Gauriau et al., 2015; Glocker et al., 2012).

The initial framework proposed by Criminisi et al. (Criminisi et al., 2009) was later advanced by other authors to include new descriptive features (Keraudren et al., 2015; Montillo et al., 2011; Pauly et al., 2011), additional shape constraints (Glocker et al., 2012), or different variations of the decision trees (Heinrich and Blendowski, 2016). Montillo et al. (Montillo et al., 2011) used of entangled trees that combine the long-range features with information of the confident voxel labels at early stages of the classification trees. Inspired by the auto-context architecture (Tu and Bai, 2010), these new features provided rich semantic context for the simultaneous location of multiple organs in abdominal CT using a single classifier. In (Glocker et al., 2012), Glocker et al. explored the use of a combined classification-regression forest for the segmentation of a set of abdominal and pelvic structures. Using distance maps as implicit shape representation, they incorporated additional structural information about the spatial arrangement of the organs and their shape, associating each image voxel to both its class label, and its distance to each organ boundaries. They also proposed a new definition of joint entropy to increase class and spatial consistency.

The feature space was also expanded by Pauly et al. (Pauly et al., 2011) who used regression forests to localize organs in multi-channel MR Dixon sequences of the whole body. They used 3D local binary patterns to capture multi-scale texture information, which provide relative intensity values and suffer field inhomogeneities. The use of binary descriptors was also explored by Heinrich and Blendowski (Heinrich and Blendowski, 2016). In this work, the authors also introduced a new RF-like classifier specifically optimized for high-dimensional feature spaces, named vantage point forest. Finally, Keraudren et al. (Keraudren et al., 2015) proposed a RF-based framework for the automatic location of multiple fetal organs using steerable features. These features were defined in a local coordinate system specific to the anatomy of the fetus to cope with its unknown orientation.

*3.6.2 Multi-Classifier Models*

The concatenation of multiple classifiers has also been an efficient strategy for the characterization and integration of contextual and high-level information into machine learning-based approaches (Gao et al., 2016; Iglesias et al., 2011; Seifert et al., 2009; Selver, 2014; Tu and Bai, 2010). In (Seifert et al., 2009), Seifert et al. proposed one of the first frameworks in which the organs were detected and segmented with a sequence of learned classifiers using marginal space learning. Starting with a simple classifier with only a few parameters (e.g., only position, not including orientation and scale), the complexity of the predictions increased as moving forward in the sequence of classifiers. They used previous predictions to obtain more refined and accurate results (e.g., full segmentation of the target organs) throughout the body. Similarly, the auto-context architecture for brain segmentation proposed by Tu and Bai (Tu and Bai, 2010) is a determinist procedure for the computation of the marginal distribution, in which the predictions of the first classifier were used as input predictive features for the next one. This approach integrated rich image appearance models, extracted from local image patches, with the anatomical context information from a series of classifiers. The auto-context configuration was also explored by Gao et al. (Gao et al., 2016) as a strategy to incorporate structural priors into a multi-organ deformable model of the pelvic region.

In (Gauriau et al., 2015), Gauriau et al. presented an alternative strategy based on the concatenation of different classifiers. In the proposed framework, a first RF-based classifier encoded global inter-organ relations, learning simultaneously the location of all the organs. Each organ was later refined combining organ-specific classifiers together with shape priors. Despite the additional computational cost associated with the training of multiple classifiers, the high computing capabilities available nowadays, together with the new and more sophisticated training strategies (Iglesias et al., 2011), make possible the use of complex and sophisticated configurations while keeping a reasonable training time.

### 3.6.3 Deep Learning Models

The development of efficient training techniques, along with increased computational power and the availability of large training data, have been particularly relevant in the adoption of deep learning (DL) approaches at the core of the machine learning research community in the last years. Naturally, the medical image analysis community has fast adopted the significant advances in this area (He et al., 2016; Krizhevsky et al., 2012; Simonyan and Zisserman, 2014; Szegedy et al., 2015), (Litjens et al., 2017), including the analysis of multiple organs.

The main limitations of traditional machine learning-based strategies are the dependency on handcrafted features, and the ability of these features to model complex inter- and intra-organ relations (e.g., the long-range spatial features). DL models are based on deep architectures composed by many layers that learn automatically features that optimally represent the data for the problem at hand, including complex multi-organ interactions. However, the big amount of manually annotated data required to train these models can still be an important limitation in the medical imaging field (see Section 4.2).

Shin et al. (Shin et al., 2013) proposed one of the first DL-based approaches for the detection of multiple organs in 3D dynamic MRI of the abdomen. The authors used a stacked sparse auto-encoder architecture to learn unsupervised temporal and spatial features from 2D image patches. These features were later used as input of the final classification network in order to generate probabilistic maps for each organ. Roth et al. (Roth et al., 2015) were among the first to explore the use of deep convolutional neural networks (CNN) for the recognition of different anatomical regions from 2D CT slices. Dividing the body into five big anatomical sections (neck, lung, liver, pelvis and legs), the proposed model assigned a region label to each image slice. However, due to its slice-by-slice nature, the model did not embed a comprehensive multi-organ anatomical knowledge effective to detect more detailed anatomical structures. Later, Yan et al. (Yan et al., 2016) used a multi-stage CNN-based framework to identify twelve body regions. Similarly to the work in (Shin et al., 2013), a first stage identified the most discriminative and informative patches then used by a second classifier for image classification.

Unlike the previous works only focused on the identification of body regions, Shakari et al. (Shakeri et al., 2016) proposed a 2D-CNN scheme for the segmentation of subcortical brain structures. To impose volumetric homogeneity on the initial slice-based segmentations, they constructed a 3D conditional random field on top of the CNN, using the output of the first network as unary potentials of a multi-label energy minimization problem. Also on the brain, the method presented by Moeskops et al. (Moeskops et al., 2016) for multi-region brain segmentation proposed the combination of multiple 2D patches at multiple resolutions as a method to obtain accurate segmentation details as well as spatial consistency: multi-organ spatial information was provided by patches at larger scales, while small scales provided detailed local information.

To alleviate the inherent limitations of 2D-based approaches (e.g., missing anatomical context in the direction orthogonal to the image plane), de Brebisson and Montana (de Brébisson and Montana, 2015) presented a hybrid framework that combined small 3D patches with larger 2.5D patches that provided a broader anatomical context. In particular, the 2.5D patches consisted of a stack of three 2D patches extracted from the three orthogonal planes of brain MRI volume. Similarly, the method presented by de Vos et al. (de Vos et al., 2017) approached the 3D localization problem as a 2D detection task, combining the output for all axial, coronal and sagittal slices to generate 3D bounding boxes for different thoracic and abdominal organs. Also in the abdomen, Wang et al. (Wang et al., 2018), and Zhou et al. (Zhou et al., 2017) proposed similar segmentation frameworks processing 2D slices of each view independently, using structural similarity (Wang et al., 2018), or simple majority voting (Zhou et al., 2017), to fuse the segmentation obtained from each orthogonal axis.

Purely 3D DL architectures have been introduced in recent years (Çiçek et al., 2016). The natural advantage of considering 3D information in a multi-organ segmentation problem was extensively validated by Milletari et al. (Milletari et al., 2017) who analyzed the impact of the amount of training data and data dimensionality (i.e., 2D, 2.5D, and 3D) for different network architectures. Dolz et al. (Dolz et al., 2018) proposed the use of a genuine 3D fully convolutional architecture for the segmentation of multiple subcortical structures of the brain. To address the computational limitations of a purely 3D approach, the authors used smaller convolutional kernels and deeper architectures (a strategy that has proven effective in the literature (He et al., 2016; Simonyan and Zisserman, 2014; Szegedy et al., 2015)). Similarly to the popular U-net architecture (Çiçek et al., 2016; Ronneberger et al., 2015; Roth et al., 2018a), Dolz et al. connected the intermediate layers' output with the final prediction layers to combine local inter-organ information with global multi-organ context into the final segmentation results. The extensive validation of the method demonstrated the great potential of DL-based models to perform complex tasks very fast. The use of 3D fully CNN was also explored by Gibson et al. (Gibson et al., 2018) and Roth et al. (Roth et al., 2018b, 2017) for the detection and segmentation of abdominal organs in CT images. In (Gibson et al., 2018), the authors combined a dense fully convolutional V-Network with spatial priors. On the other hand, Roth et al. (Roth et al., 2018b, 2017) presented a coarse-to-fine strategy in two-stages: a first network was trained to provide a rough delineation of each organ, learning global anatomical information and inter-organ relations; using these results as input, a second network focused on local anatomical details, providing a more detailed segmentation of each organ.

The concatenation of multiple networks was also explored by Wachinger et al. (Wachinger et al., 2018) to mitigate the class imbalance problem, though only the foreground and background were identified by the first network. In the second network, they used a 3D patch-based approach, including the coordinates of the central voxel to mitigate the lack of spatial context when working with patches. Similarly to the framework presented in (Shakeri et al., 2016), the final segmentation was generated by a 3D fully connected conditional random field, which ensured label agreement between voxels and imposed spatial and volumetric consistency to the organs. Finally, Hu et al. (Hu et al., 2017) also refined the initial segmentation provided by a fully 3D CNN by using the resulting probabilistic maps as initialization for a subsequent fine segmentation based on level sets. The authors used the multi-organ prediction maps generated by the first CNN as additional spatial constraints, together with the traditional intensity-based models and disjoint regions constraints. A two stage cascaded architecture was also used by Chen et al. (Chen et al., 2017b) for the segmentation of four abdominal organs (liver, spleen left and right kidney) in dual energy CT images. In particular, the authors concatenated two U-net-like fully convolutional networks, using the region of interest generated by the first one to reduce the search space for the second stage. Finally, a similar approach was recently proposed by Valindria et al. (Valindria et al., 2018), combining a two-stage convolutional network with spatial atlas priors in order to improve the segmentation accuracy of small organs in whole-body MRI scans. These patient-specific anatomical priors have proven particularly useful to deal with the large variability of abdominal organs.

One of the most promising trends in the application of DL-based methods to the analysis of multiple anatomical organs is the synergy between the high predictive power of these deep networks and the anatomical constraints provided by single- or multi-organ shape models. In this line, recent works (Mansoor et al., 2017; Milletari et al., 2017; Oktay et al., 2017) presented interesting alternatives to the most common state-of-the-art methods that operate as simple voxel-based classifiers. Recently, Mansoor et al. (Mansoor et al., 2018, 2017) proposed a marginal shape DL model for the segmentation of thoracic radiographs, encoding a multi-organ statistical models as an image-based classification problem. On the other hand, Milletari et al. (Milletari et al., 2017) presented a combination of CNN-based classifier with a Hough voting strategy that encodes shape constraints in the segmentation of deep brain regions. As demonstrated by these works, the integration of shape priors into DL-based architectures provides smooth and anatomically consistent results, even under the constraints of limited training data and computational resources.

## 3.7 *Graph-Based Models*

The graph-based representation of models is intuitive and compact, and has the ability to model complex probabilistic systems. Graph-based models were at the foundation of some of the most popular segmentation techniques (Paragios et al., 2016), including probabilistic MRF (Kato and Pong, 2006) and Bayesian networks (Zhang and Ji, 2010)), and deterministic graphical models (e.g., graph cuts (Boykov and Funka-Lea, 2006)). Naturally, these techniques were first used for single-organ analysis and were later extended to more general multi-label applications (Wang et al., 2009). Graph-based segmentation models map the entire image on a graph where the nodes correspond to low-level structural elements (e.g., pixels or super-pixels), and the neighborhood is expressed by edges connecting the nodes. This bottom-up structuring of the information represents a significant difference from the graph-based representation used by the sequential models discussed in Section 3.4.1 to establish the order in which each organ is processed. For example, in these graph-based models, the segmentation task is considered as an energy function minimization problem in which all the organs are simultaneously segmented via efficient optimization algorithms (e.g., graph-cut).

The segmentation of multiple organs using standard graph-based approaches has been successfully explored for the abdominal and thoracic region (Bajger et al., 2013; Dong et al., 2016; Linguraru and Summers, 2014). However, the bottom-up approach used in these models presents an important limitation: the need for additional constraints to ensure the anatomical consistency of the results in a complex multi-organ scenario. An effective strategy to bring high-level information into graph-based models has been the incorporation of probabilistic atlases (Freedman and Zhang, 2005; Linguraru et al., 2012; Park et al., 2010; Song et al., 2006). Bhole et al. (Bhole et al., 2014) used Gaussian mixture models (GMMs) to encode high level semantic information into different graph-based models, including MRFs, Conditional random fields (CRFs) and kernel CRFs.. In (Linguraru et al., 2012), Linguraru et al. used abdominal probabilistic atlases and Parzen shape windows (Parzen, 1962) to encode, respectively, spatial and shape constraints directly into the 4D graph-cut-based segmentation framework, as new terms of the energy function. Alternatively to probabilistic atlases, Bagci et al. (Bagci et al., 2012) used PCA-based shape atlases to initialize a multi-organ graph-cut segmentation by establishing relation functions of shape and intensity appearance patterns. Finally, Kéchichian et al. (Kechichian et al., 2013; Kéchichian et al., 2014) proposed an interesting combination of graph-cut with higher-level (i.e., organ-based) graph models to impose inter-organ vicinity priors. These priors were defined as the shortest-path pairwise constraints on a graph model of inter-object adjacency relations to define the order of analysis, similarly to those used in sequential models (Section 3.4).

*3.8 Articulated Models*

In the majority of the works reviewed in the paper, the variations in the relative position between organs were modeled based on the inherent population variability. Such approaches may be appropriate for non-articulated structures, such as the abdomen or the brain, but less relevant for articulated musculoskeletal structures. The analysis of the relative position among elements in a sequence of multi-postural images, and/or their motion, may provide the most valuable clinical information for treatment and diagnosis (Delp et al., 1990; Kuo et al., 2009; Liu et al., 2008; Smoger et al., 2015).

Articulated models focus on the characterization of the inter-object spatial arrangement of the musculoskeletal complex, where inter-object relative position are usually described as a collection of pairwise rigid transformations. These models can be grouped into two main categories: kinematically-constrained registration models, where patient-specific deformations are treated as a registration problem; and statistical articulated models, in which inter-organs relations are assimilated by the model as part of the inherent variability of the multi-organ complex.

*3.8.1  Kinematically-Constrained Registration Models*

While bones joints can be described using rigid transformations, the surrounding tissues deform in a more complex way, making image-based registration approaches challenging (du Bois d'Aische et al., 2007; Li et al., 2006). Early works on articulated models incorporated a segmentation stage of the bony structures of interest in all the images of a multi-postural sequence to help the convergence of the subsequent registration process (Kamojima and Miyata, 2004; Liu et al., 2008). Other authors used simplified geometrical representations of the articulated complex (Bois et al., 2005; Chen et al., 2010; Martin-Fernandez et al., 2009; Miyata et al., 2003). One of the first models (Bois et al., 2005) used a kinematic model of the vertebrae (Monheit and Badler, 1991) represented as a set of simple geometrical structures connected by nodes; deformation constraints were explicitly imposed between pairs of elements. A similar simplified geometrical model was used by Miyata et al. (Miyata et al., 2003) to define centers and rotation axes of the joints of the hand, also inspiring Martin et al. (Martin-

Fernandez et al., 2009) to use a wire model of the inner bone skeleton of the hand in a poly-affine-based articulated registration framework. A more detailed model of the bones of the hand was presented by Chen et al. (Chen et al., 2010). They used a complex hierarchical structure of 20 segments, including the capitate bone of the wrist, metacarpals, and phalanges. The reference model also incorporated information of the finger joints (e.g., rotation ranges) to constrain the motion of the bones. Finally, Gill et al. (Gill et al., 2012) proposed a multi-modal ultrasound-CT registration method of the lumbar spine, which explicitly incorporated biomechanical constraints (Desroches et al., 2007; Panjabi et al., 1976) (i.e., relations between the displacement of the intervertebral structures, reaction forces, and moments) to guarantee plausible instances of the model from an anatomical and mechanical point of view.

### 3.8.2 Statistical Articulated Models

PDMs are the foundation of statistical articulated models. Despite their versatility and flexibility, linear PDMs have limitations to efficiently characterize the non-linearity of joint movements, such as rotational movements in the spine (Ali et al., 2012), hip (Yokota et al., 2009), or knee (Smoger et al., 2015). The use of non-linear models, such as polynomial regression PDM (Sozou et al., 1994), combined cartesian-polar PDM (Heap and Hogg, 1996), multi-layer perceptrons (Sozou et al., 1997), or kernel PCA (Twining and Taylor, 2001), allows for a more natural and accurate representation of rotations. These techniques have been commonly used in computer vision applications to represent human-body movements (Al-Shaher and Hancock, 2004; Bowden, 2000).

To model articulations in multi-organ analysis, Kainmüeller et al. (Kainmüeller et al., 2009a) and Balestra et al. (Balestra et al., 2014) proposed an extension of classic PCA-based shape model by incorporating a rotation matrix that explicitly described the relative transformation of the moving organs. Such representations used simplified anatomical models (e.g., approximating the hip joint by an ideal ball-and-socket joint (Kainmüeller et al., 2009a)) with a restricted range of movements. To overcome this limitation, other authors (Boisvert et al., 2008a, 2006) used frames (i.e., points associated with three orthogonal axes) and Riemannian geometry to model relative orientations and positions. For example, Boisvert et al. (Boisvert et al., 2006) proposed a hybrid model of the vertebral body that combined a global statistical model of the Riemannian manifold of inter-vertebrae rigid transformations with local vertebral shape models. A number of other works on the spine (Boisvert et al., 2008b; Harmouche et al., 2012; Klinder et al., 2008; Moura et al., 2011; Rasoulian et al., 2013), and the wrist (Anas et al., 2016; Chen et al., 2014), were also inspired by this hybrid model.

As an alternative to Riemannian manifolds, Kadoury and Paratios (Kadoury and Paragios, 2009) proposed an articulate model using MRF graphs to characterize the relative intervertebral transformations. Finally, Constantinescu et al. (Constantinescu et al., 2016) combined statistical shape models with finite element analysis to artificially enlarge the number of samples needed to effectively characterize the complex movement of the knee. Similarly to Boisvert et al. (Boisvert et al., 2006), the multi-pose shape model in (Constantinescu et al., 2016) was built by applying PCA on the manifold of the parameters of the rigid inter-organ transformations, a strategy that has been proven to be more effective than operating on explicit contours.

## 4. Discussion

Our review of computational anatomy techniques from medical images has shown that methodological strategies such as multi-level-based analysis, and sequential models, are particularly suitable for multi-organ modeling and analysis. However, recent advances in artificial intelligence have introduced efficient and flexible machine learning techniques applied to the analysis of multiple organs. Machine learning will continue to play a fundamental role in the development of future multi-organ analysis frameworks, likely in combination with other techniques, such as the aforementioned multi-level or sequential models. Another general observation worth highlighting is the prevalence of region-specific models, particularly the brain and the abdomen. In this section, we discuss the applications of multi-organ analysis techniques to anatomical regions, as well as the limitations and opportunities in the growing field of computational anatomy.

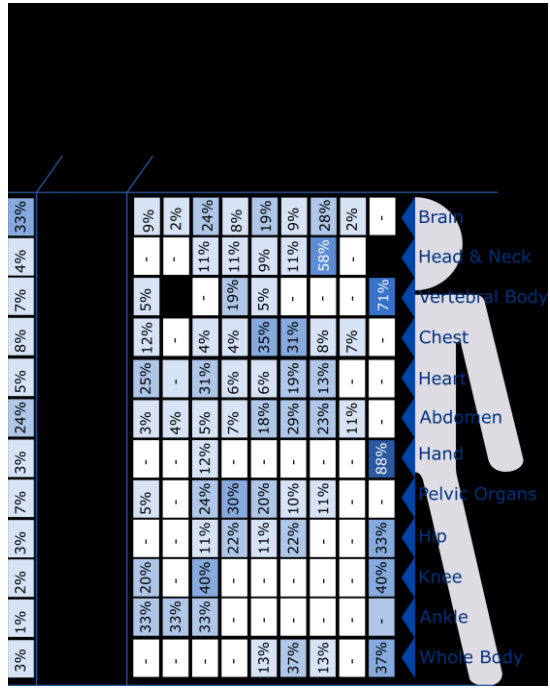

Fig. 3. Distribution by anatomical region of the papers included in this review. The figure also includes the average, median, and maximum number of structures analyzed in each anatomical area. On the right side, the figure also shows the most common analysis techniques for each anatomical region.

## 4.1 Analysis by Anatomical Region

Multi-organ analysis techniques have been applied to most anatomical regions of the human body, from the semantic segmentation of subcortical brain structures at the top (Cerrolaza et al., 2012) to motion analysis of the ankle complex at the bottom (Liu et al., 2008). Fig. 3 shows the distribution by anatomical region of the more than 300 papers included in this survey. The figure also includes information about the techniques most commonly used for each region, as well as statistics on the number of structures. See Table I for a detailed list of the publications reviewed in this paper organized by anatomical region and analysis technique.

Together with the heart, the brain and the abdomen are the regions that have traditionally been most studies by the medical image analysis community. A similar trend is observed in the context of multi-organ analysis, with the brain being the subject of study in 33% of the works reviewed here. The study of the human brain as a modular structure ranged from the simpler gray/white matter division of the brain tissue (Merino-Caviedes et al., 2010; Song et al., 2006; Zeng et al., 1999) to the more detailed partition into dozens of structures (Batrancourt et al., 2006; Heckemann et al., 2006; Lancaster et al., 1997; Milletari et al., 2017). The increasing resolution of brain imaging technology offering detailed and accurate representation of this organ also encouraged the development of new computational approaches to characterize multiple neuro-anatomical structures. These approaches have a wide-range of clinical applications, such as the characterization of the brain morphology associated with specific pathologies (e.g., Alzheimer's disease, or autism (van Rooij et al., 2018)). For brain analysis, surface-based morphometric analysis (Wade et al., 2015) has been preferred over voxel-based methods for two main reasons: they provide accurate analysis of the multi-structural composition of the brain, and they infer anatomical meaning to the analysis. Most of the morphometric studies rely on registration techniques to warp the brain to a reference space and compare structural difference between patients. But as shown in Fig. 3, brain analysis has benefited from practically all multi-organ analysis techniques, from the more traditional approaches (e.g., global-, atlas-, and coupled deformable models-based techniques) to the more sophisticated and flexible multi-resolution and sequential frameworks that include inter-region relations (Cerrolaza et al., 2016; Rao et al., 2008; Rousson and Xu, 2006) and multi-scale analysis (Cerrolaza et al., 2015).

At the second place in order of popularity, and addressed by 24% of the papers reviewed, the analysis of the abdominal region remains one of the most challenging problems in the field of medical image analysis. Compared to the brain, the abdominal organs present a higher inter-subject variability, due to differences in age, gender, stature, or disease status, as well as more complex inter-organ relations. Additional challenges are induced by body pose, respiratory cycle, edema, or digestive status. As depicted in Fig. 3, atlas-based methods have been very popular in the study of the abdominal region. In particular, hierarchical probabilistic atlas (Chu et al., 2013; Okada et al., 2008b), or multi-atlas models with a hierarchical voting scheme (Wolz et al., 2013) generally performed better than intensity-based registration methods, which were more severely affected by the size and contrast difference between organs (e.g., liver vs. gallbladder). The use of sequential models has also proved to be particularly

Table I. List of publications per anatomical region (rows), and per principal analysis technique.

| Region | Publications |
|---|---|
| Brain | **GlM▶** Akhoundi-Asl and Soltanian-Zadeh, 2007; Bossa and Olmos, 2006; Bossa and Olmos, 2007; Bossa et al., 2011; Cootes et al., 1994; Duta and Sonka, 1998; Gorczowski et al., 2010; Poupon et al., 1998; Styner et al., 2006; **InM▶** Akhoundi-Asl and Soltanian-Zadeh, 2007; Asl and Soltanian-Zadeh, 2008; **CoDeM▶** Bazin and Pham, 2006; Fan et al., 2008a; Fan et al., 2008b; Gao et al., 2011; Gao et al., 2017; Han and Prince, 2003; Han et al., 2002; Ho and Shi, 2004; Holtzman-Gazit et al., 2003; Jeon et al., 2005; Kim et al., 2014; Litvin and Karl, 2005; MacDonald et al., 1994; Mangin et al., 1995; Merino-Caviedes et al., 2010; Pohl et al., 2007; Samson et al., 2000; Tsai et al., 2001; Uzunbas et al., 2010; Vese and Chan, 2002; Wang et al., 2013; Yan et al., 2009; Yang et al., 2004; Zeng et al., 1999; **MuLeM▶** Cerrolaza et al., 2011; Cerrolaza et al., 2012; Cerrolaza et al., 2014; Cerrolaza et al., 2015; Cerrolaza et al., 2016; Joohwi Leea, Sun Hyung Kimb, 2016; Shen et al., 2001; Zhang et al., 2011; **SeqM▶** Atif et al. 2007; Batrancourt et al., 2006; Bloch et al. 2003; Bloch et al., 2005; Cerrolaza et al., 2016; Colliot et al., 2006; Fouquier et al., 2007; Fouquier et al., 2008; Fouquier et al., 2012; Géraud et al., 1999; Holtzman-Gazit et al., 2003; Hudelot et al., 2008; Jeon et al., 2005; Puentes et al., 2008; Rao et al., 2006; Rao et al., 2008; Rousson et al., 2006; Sofka et al., 2010; Udupa and Saha, 2003; **MaLrM▶** Dolz et al., 2018; Fischer et al., 2014; Milletari et al., 2017; Moeskops et al., 2016; Shakeri et al., 2016; Sofka et al., 2010; Tu and Bai, 2010; Wachinger et al., 2018; de Brébisson and Montana, 2015; **AtM▶** Aljabar et al., 2009; Bazin and Pham, 2008; Chiavaras et al., 2001; Cocosco et al., 2003; Collins et al., 1995; Dawant et al., 1999; Evans et al., 1991; Fischl et al., 2002; Gholipour et al., 2017; Géraud et al., 1999; Han and Fischl, 2007; Heckemann et al., 2006; Joohwi Leea, Sun Hyung Kimb, 2016; Klein et al., 2005; Lancaster et al., 1997; Mazziotta et al., 1995; Mazziotta et al., 2001; Pohl et al., 2006; Rohlfing et al., 2003a; Rohlfing et al., 2003b; Rohlfing et al., 2004; Sandor and Leaby, 1997; Talairach and Tournoux, 1988; Thompson et al., 2000; Van Leemput et al., 1999; Wang et al., 2013; Warfield et al., 2004; van Buren and Maccubbin, 1962; **GrM▶** Kechichian et al., 2013; Song et al., 2006; |
| Head & Neck | **CoDeM▶** Kainmueller et al., 2009; **MuLeM▶** Merck et al., 2008; **SeqM▶** Udupa et al., 2013; **MaLrM▶** Pauly et al., 2011; **AtM▶** Chen et al., 2017a; Chen et al., 2017a; Fritscher et al., 2014; Fritscher et al., 2014; Han et al., 2008; Han et al., 2008; |
| Vertebral Body | **GlM▶** Smyth et al., 1997; **MuLeM▶** Bernard et al., 2001; Lecron et al., 2012a; Lecron et al., 2012b; Neubert et al., 2014; **SeqM▶** de Bruijne et al. 2007; **ArtM▶** Ali et al., 2012; Bois et al., 2005; Boisvert et al., 2006; Boisvert et al., 2008a; Boisvert et al., 2008b; Desroches et al., 2007; Gill et al., 2012; Harmouche et al., 2012; Kadoury and Paragios, 2009; Klinder, 2008; Monheit and Badler, 1991; Moura et al., 2011; Panjabi et al., 1976; Rasoulian et al., 2013; du Bois d'Aische et al., 2007; |
| Chest | **GlM▶** Mansoor et al., 2017; Mansoor et al., 2018; van Ginneken et al., 2006; **CoDeM▶** Brock et al., 2005; **MuLeM▶** Cerrolaza et al., 2011; **SeqM▶** Camara et al., 2004; He et al., 2015; Kéchichian et al., 2014; Matsumoto and Udupa, 2013; Sun et al., 2016; Udupa et al., 2011; Udupa et al., 2013; Wang et al., 2014a; Wang et al., 2014b; **MaLrM▶** Criminisi et al., 2009; Iglesias et al., 2011; Keraudren et al., 2015; Mansoor et al., 2017; Mansoor et al., 2018; Montillo et al., 2011; Pauly et al., 2011; de Vos et al., 2017; **AtM▶** Gubern-Mérida et al., 2011; Schreibmann et al., 2014; **GrM▶** Bajger et al., 2013; Kéchichian et al., 2014; |
| Heart | **GlM▶** Cootes et al., 1994; Cootes et al., 1995; Frangi et al., 2002; Schwarz et al., 2010; **CoDeM▶** Arrieta et al. 2017; Gao et al., 2017; Kohlberger et al., 2007; Schwarz et al., 2010; Uzumbas et al., 2013; **MuLeM▶** Hui Sun et al., 2008; **SeqM▶** Sofka et al., 2010; **MaLrM▶** Keraudren et al., 2015; Seifert et al., 2009; Sofka et al., 2010; **AtM▶** Frangi et al., 2002; Isgum et al., 2009; |
| Abdomen | **GlM▶** Bagci et al., 2012; Gollmer et al., 2012; **InM▶** Bagci et al., 2012; Gollmer et al., 2012; Yao and Summers, 2009; **CoDeM▶** Brock et al., 2005; Gao et al., 2011; Gao et al., 2017; Kohlberger et al., 2011; **MuLeM▶** Cerrolaza et al., 2015; Cerrolaza et al., 2016; Okada et al., 2008b; Valenzuela et al., 2015; Vera et al., 2012; **SeqM▶** Camara et al., 2004; Cerrolaza et al., 2016; Fasquel et al., 2006; He et al., 2015; Huang et al., 2014; Kéchichian et al., 2014; Okada et al 2013; Okada et al., 2012; Okada et al., 2015; Shimizu et al., 2010; Udupa et al., 2013; Wang et al., 2014a; Wang et al., 2014b; Wang et al., 2015; **MaLrM▶** Chen et al. 2017b; Criminisi et al., 2009; Criminisi et al., 2013; Fischer et al., 2014; Gauriau et al., 2015; Gibson et al., 2018; Glocker et al., 2012; Heinrich and Blendowski, 2016; Hu et al., 2017; Iglesias et al., 2011; Keraudren et al., 2015; Montillo et al., 2011; Pauly et al., 2011; Roth et al., 2017; Roth et al., 2018a; Roth et al., 2018b; Seifert et al., 2009; Selver, 2014; Shin et al., 2013; Wang et al., 2018; Zhou et al., 2017; de Vos et al., 2017; **AtM▶** Chu et al., 2013; Linguraru et al., 2009; Linguraru et al., 2010; Liu et al., 2010; Oda et al., 2012; Okada et al 2013; Okada et al., 2008b; Okada et al., 2012; Park et al., 2003; Park et al., 2010; Schreibmann et al., 2014; Shimizu et al., 2010; Suzuki et al., 2012a; Suzuki et al., 2012b; Wolz et al., 2013; Xu et al., 2015; Zhou et al., 2007; **GrM▶** Bajger et al., 2013; Bhole et al., 2014; Dong et al., 2016; Kechichian et al., 2013; Kéchichian et al., 2014; Linguraru and Summers, 2014; Linguraru et al., 2012; Oda et al., 2012; |
| Hand | **CoDeM▶** Han and Prince, 2003; **ArtM▶** Anas et al., 2016; Chen et al. ,2014; Chen et al. 2010; Kamojima and Miyata, 2004; Kuo et al., 2009; Martin-Fernandez et al., 2009; Miyata et al., 2003 |
| Pelvis | **GlM▶** Li et al., 2016; **CoDeM▶** Costa et al., 2007; Hensel et al., 2007; Ma et al., 2013; Rousson et al., 2005; Song et al., 2013; **MuLeM▶** Chaney et al., 2013; Fletcher et al., 2002; Jeong et al., 2006; Lu et al., 2007; Merck et al., 2008; Pizer et al., 2005; **SeqM▶** Chandra et al., 2016; Jeong et al., 2008; Ma et al., 2010; Pizer et al., 2005; **MaLrM▶** Gao et al., 2016; Seifert et al., 2009; **AtM▶** Akhondi-Asl et al., 2014; Weisenfeld and Warfield, 2011; |
| Hip | **CoDeM▶** Kainmüeller et al., 2009b; **MuLeM▶** Bukovec et al., 2011; Yokota et al., 2009; **SeqM▶** Yokota et al., 2013; **MaLrM▶** Glocker et al., 2012; Montillo et al., 2011; **ArtM▶** Balestra et al., 2014; Kainmüeller et al., 2009a; Yokota et al., 2009; |
| Knee | **GlM▶** Fripp et al., 2007; **CoDeM▶** Pang et al., 2015; Uzumbas et al., 2013; **ArtM▶** Constantinescu et al., 2016; Smoger et al., 2015; |
| Ankle | **GlM▶** Bagci et al., 2012; **InM▶** Bagci et al., 2012; **CoDeM▶** Li et al., 2006; |
| Whole Body | **SeqM▶** Zhan et al., 2008; **MaLrM▶** Roth et al., 2015; Valindria et al., 2018; Yan et al., 2016; **AtM▶** Wang et al., 2012; **ArtM▶** Al-Shaher and Hancock, 2004; Bowden, 2000; Li et al., 2006; |

**GlM▶** Global Models; **InM▶** Individual Models; **CoDeM▶** Coupled Def. Models; **MuLeM▶** Multi-Level Models; **SeqM▶** Sequential Models; **MaLrM▶** Machine Learning Models; **AtM▶** Atlas Models; **GrM▶** Sequential Models; **ArM▶** Sequential Model effective in the abdominal region. Such models allowed to implicitly incorporate the anatomical relations between organs, and thus improved the accuracy of the analysis of the most challenging and variable organs (e.g., the pancreas or gallbladder). The definition of constraints and contextual information provided by the more stable surrounding organs (e.g., liver, spleen, and kidneys) has provide an evident benefit for the creation of computational anatomical models of the abdomen (Cerrolaza et al., 2016; Huang et al., 2014; Okada et al., 2012; Wang and Smedby, 2014a). Most recently, DL-based techniques have shown promising results in the simultaneous location of multiple abdominal organs (Litjens et al., 2017; Roth et al., 2015; Shin et al., 2013). However, DL methods for multi-organ segmentation are still limited to 2D and patch-based models by the large size of the abdominal scans (which increases the computational cost of native 3D and fully-connected architectures).

The chest, the vertebral body, and the pelvic region have also been popular with multi-organ analysis techniques, appearing in 8%, 7%, and 7% of the papers, respectively. Similarly to the abdominal region, sequential (Sun et al., 2016; Udupa et al., 2011) and machine learning-based methods (Keraudren et al., 2015; Mansoor et al., 2017) have been applied to the study of the chest, with many papers tackling the abdomen and thorax within the same framework (Camara et al., 2004; Criminisi et al., 2009; de Vos et al., 2017; Iglesias et al., 2011; Montillo et al., 2011; Wang and Smedby, 2014b). For the analysis of the vertebral body, its inherent multi-structural composition inspired the use multi-resolution models, with the vertebrae representing the elementary structural level (Lecron et al., 2012a, 2012b; Neubert et al., 2014; Pereañez et al., 2015), and articulated methods, with explicit modeling of pairwise inter-vertebrae deformation and mechanical properties (Desroches et al., 2007; Gill et al., 2012; Monheit and Badler, 1991; Panjabi et al., 1976).

The analysis of the pelvic organs (7% of the papers) comprised the study of the bladder, rectum, and prostate in the case of the male pelvis, and vagina in the female case using a variety of techniques from the entire methodological spectrum. The other anatomical regions, outside of the brain, abdomen, chest, pelvis and vertebral body, have received comparatively marginal attention (e.g., only 4% of the papers analyzed the head and the neck, 3% the hand or the hip, and 2% and 1% the knee and the ankle, respectively). The approaches used to analyze these regions have also been less varied and region-specific (e.g., articulated models of the hand or ankle, and atlas-based models of the head and neck).

Finally, as shown in Fig. 3 and Table I, only a few works (~3%) have addressed the analysis of the whole body, using sequential, atlas, machine learning, and articulated models. One of the main limitations in the construction of more holistic and global anatomical models is the lack of large data sets to characterize the complexity of the human anatomy. This and other limitations are discussed below.

## 4.2 Limitations, Challenges and Future Trends

The limited availability of annotated clinical data has been a recurrent problem in the medical imaging field. This limitation becomes particularly relevant when modeling multi-organ anatomical structures, where large datasets are needed to characterize not only the particular locality of each organ, but also the complex inter-organ relations. The need for large datasets is therefore vital for the development and validation of new multi-organ analysis approaches, and represents a major obstacle to realize the full potential of DL-based techniques. Although publicly available databases with manual annotations of multiple anatomical structures exist, the number of cases is usually limited to a few dozen to a couple of hundred volumes at best. Popular data repositories include the Internet Brain Segmentation Repository (IBSR, n.d.) and the Open Access Series of Imaging Studies (OASIS, n.d.) brain databases, which provide open access to detailed manually-guided expert volumetric segmentations of subcortical structures in MR brain data; the VISCERAL dataset (Jimenez-del-Toro et al., 2016) includes up to 18 annotated anatomical structures from the chest, abdomen, and pelvis; or the Challenge on Spine Imaging 2014 (CSI'14) vertebra segmentation database (Yao et al., 2016) with manual annotations of thoracic and lumbar vertebrae in CT volumes.

Recent large initiatives, such as the NIH Cancer Imaging Archive (Clark et al., 2013) and the UK Biobank Imaging Study (Sudlow et al., 2015), provide open access to a comprehensive database with thousands of images, including CT, MRI, PET and more images of the heart, abdomen, brain and bones, among others. Annotations are not available on most of these images, but the combination of datasets includes a mix of healthy and pathological cases, which is critical to develop future computational anatomy methods and CAD systems that are robust to pathology and unusual anatomy (e.g., accurate abdominal segmentation of patients with pancreas, or liver cancer; or the analysis of inter-organ relations when one of the organs has been removed (Suzuki et al., 2012a, 2012b)).

In this context, the development of new image-based diagnostic tools has been strongly influenced by machine learning techniques, and specially the new DL-based architectures. But these machine learning techniques have their own challenges (see Section 3.6.3). We have already mentioned the critical need for large datasets to train these heavily over-parameterized models. It is also important to consider that unlike the "black-box" nature of DL models, the integration of inter- and intra-organ statistical models in the form of shape constraints could provide anatomically meaningful architectures that guarantee the coherence of the results, while also reducing training data requirements (Mansoor et al., 2018, 2017; Milletari et al., 2017; Oktay et al., 2017). On the other hand, the high computational cost associated to these networks usually hampers the efficient use of the progressively higher-resolution new imaging modalities. As already mentioned in this review, strategies to reduce the complexity and cost have been adopted, using patch-based, or region-specific models. But the full potential of machine learning for computational anatomy will be unlocked with the creation of new efficient architectures that operate with large amounts of image-based information as well as anatomical and physiological context. These models will decisively contribute to solving major challenges in the medical image field: from the creation of accurate computer-aided diagnosis systems that integrate inter-organ-based features, to patient specific surgery planning (Morimoto et al., 2017).

As mentioned in the previous section (Section 4.1), the development of comprehensive and global computational models of the human body is still one of the great challenges in computational anatomy. These models could benefit from the integration of multiple imaging modalities that allow to characterize inter-organ relations, as well as organ-specific mechanical properties (e.g., tissue properties, stress and strain) at different scales. On the other hand, another important limitation not yet addressed in multi-organ analysis is the inherent inter- and intra-organ variability with body development and age. The creation of more comprehensive and anatomically accurate multi-organ models able to effectively characterize the complexity of the human anatomy, its diversity, and change with age (even during the embryotic (de Bakker et al., 2016) and fetal stage (Gholipour et al., 2017)), represents one of the biggest challenges and opportunities for the future of computational anatomy and the next generation of CAD systems.

## 5. Conclusions

In this paper, we have presented the first detailed review of computational anatomy methods for image-based multi-organ analysis. The collection and dissemination of large medical image databases, together with the continuing progress of computing capabilities, have favored the development of complex and comprehensive anatomical models. As shown in this survey of over 300 publications, the automatic analysis of multi-organ anatomical complexes has been approached from different perspectives (e.g. detection, segmentation, diagnosis, inter-organ relations), using a variety of methodologies (e.g. anatomical atlases, shape models, graphs, machine learning), and has been applied to multiple anatomical regions (e.g. brain, abdomen, pelvis, chest). The categorization of approaches in this paper provides a reference guide to the current techniques available for the analysis of multiple anatomical structures. We have also indicated current challenges and future opportunities in multi-organ analysis, including the creation of holistic multi-scale multi-organ and whole body models. New efficient computational and machine learning models must embed the anatomical and physiological context inherent to the human body to provide the essential architectures of a computational anatomist.

## Acknowledgements

This paper was supported in part by the Marie Skodoska-Curie Actions of the UE Framework Program for Research and Innovation, under REA grant agreement 706372.